\documentclass{article} 
\usepackage{iclr2024_conference,times}

\usepackage{amsmath,amsfonts,bm}

\def\eqref#1{equation~\ref{#1}}

\def\1{\bm{1}}

\DeclareMathAlphabet{\mathsfit}{\encodingdefault}{\sfdefault}{m}{sl}
\SetMathAlphabet{\mathsfit}{bold}{\encodingdefault}{\sfdefault}{bx}{n}

\usepackage{url}
\usepackage{graphicx}
\usepackage{amsmath}
\usepackage{amssymb}
\usepackage{booktabs}
\usepackage{comment}
\usepackage{csquotes}
\usepackage{bm}

\usepackage{subcaption} 

\usepackage{times}
\usepackage{epsfig}
\usepackage{graphicx}
\usepackage{amsmath}
\usepackage{amssymb}
\usepackage{booktabs}

\usepackage{booktabs} 
\usepackage{tabularx} 
\usepackage{array}    

\usepackage{caption}
\usepackage{subcaption}

\usepackage{dblfloatfix}
\usepackage{tikz}
\usetikzlibrary{fit}
\usepackage{nicematrix}
\usepackage{comment}
\usepackage{color,soul}

\usepackage{flushend}

\usepackage[pagebackref=true,breaklinks=true,letterpaper=true,colorlinks,bookmarks=false, citecolor=black, linkcolor=black, urlcolor=black]{hyperref}

\usepackage[capitalize]{cleveref}
\crefname{section}{Sec.}{Secs.}
\Crefname{section}{Section}{Sections}
\Crefname{table}{Table}{Tables}
\crefname{table}{Tab.}{Tabs.}

\renewcommand{\mkbegdispquote}[2]{%
  \begin{minipage}[t]{0.1\textwidth}#2\end{minipage}%
  \begin{minipage}[t]{0.915\columnwidth}%
}
\renewcommand{\mkenddispquote}[2]{\end{minipage}}

\usepackage{tabularx}
\usepackage{makecell}

\title{Select High-Level Features: Efficient Experts from a Hierarchical Classification Network}

\author{Andr\'e Kelm, Niels Hannemann\thanks{Niels Hannemann, Bruno Heberle, and Lucas Schmidt contributed equally to this work as part of their bachelor thesis.} , Bruno Heberle$^*$, Lucas Schmidt$^*$, Tim Rolff, \\\textbf{Christian Wilms, Ehsan Yaghoubi, Simone Frintrop}\\
Department of Computer Vision\\
University of Hamburg\\
Hamburg, Germany \\
{\ttfamily\small andre.kelm@, 0hannema@informatik., bruno.heberle@, lucas.schmidt-} \\ 
{\ttfamily\small 1@studium., tim.rolff@, christian.wilms@, ehsan.yaghoubi@,  simone.} \\ 
{\ttfamily\small frintrop@uni-hamburg.de} \\ 
}

\iclrfinalcopy 
\begin{document}

\maketitle

\begin{abstract}
This study introduces a novel expert generation method that dynamically reduces task and computational complexity without compromising predictive performance. 
It is based on a new hierarchical classification network topology that combines sequential processing of generic low-level features with parallelism and nesting of high-level features.
This structure allows for the innovative extraction technique: 
the ability to select only high-level features of task-relevant categories.
In certain cases, it is possible to skip almost all unneeded high-level features, which can significantly reduce the inference cost and is highly beneficial in resource-constrained conditions. We believe this  method paves the way for future network designs that are lightweight and adaptable, making them suitable for a wide range of applications, from compact edge devices to large-scale clouds. 
In terms of dynamic inference our methodology can achieve an exclusion of up to 88.7\,\% of parameters and 73.4\,\% fewer giga-multiply accumulate (GMAC) operations, analysis against comparative baselines showing an average reduction of 47.6\,\% in parameters and 5.8\,\% in GMACs across the cases we evaluated.
\end{abstract}

\section{Introduction}
\cite{pmlr-v162-wortsman22a} have shown that one way to achieve performance advantages in various deep learning (DL) tasks is to use larger models and more training material.
High computational power and multi-stage training processes \citep{Woo_2023_CVPR} create top performance, but typically do not fit easily on edge devices with limited resources.
\cite{cheng2023survey} summarize strategies such as model pruning, quantization, dynamic inference costs, and efficient architectures, highlighting their efficiency, but also noting the challenge of maintaining full performance.

Our method is intended to be orthogonal to these methods and offers the possibility of reducing the computational complexity without compromising predictive performance.
Unlike other efficient methods, 
where efficiency is often learned within the optimization process,
our approach can be controlled extrinsically to skip task-irrelevant high-level features and directly incorporate the essentials: e.g., the super-category of marine animals to detect sharks, or the super-category of plants to detect corn varieties.
These task-oriented configurations are called 'experts'.

\section{Related Works}
Hierarchical networks \citep{HCNN} are gaining interest as \cite{10.1007/978-3-030-93247-3_18} showed their ability to improve model performance and efficiency. 
There are two key differences from other hierarchical networks: 1) Branches are not connected by fully connected (FC) layers or other operations in a later step, only by softmax! 2) Nesting is supported by a large language model. 
To the best of our knowledge, our approach of generating efficient experts by selecting high-level features is unique.

\section{Method}
\setlength{\fboxrule}{0.5px} 
\setlength{\fboxsep}{0pt}  
\begin{figure}
  \begin{subfigure}[b]{0.3\textwidth}
  \fbox{%
    \includegraphics[clip, trim=19.7cm 19.7cm 19.7cm 19.7cm, width=\textwidth, ]    {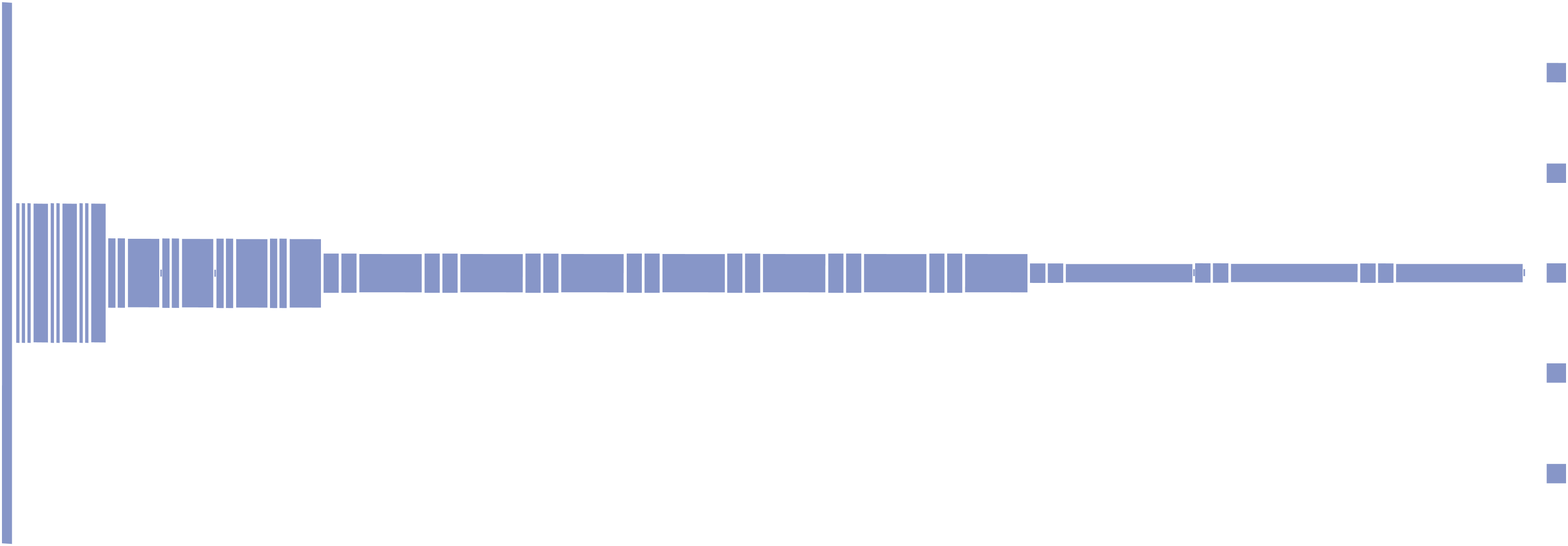}}
    \caption{Conventional}
    \label{fig:1a}
  \end{subfigure}
  \hfill
  \begin{subfigure}[b]{0.3\textwidth}
    \fbox{%
    \reflectbox{\includegraphics[clip, trim=19.7cm 19.7cm 19.7cm 19.7cm, width=\textwidth, angle=180]   {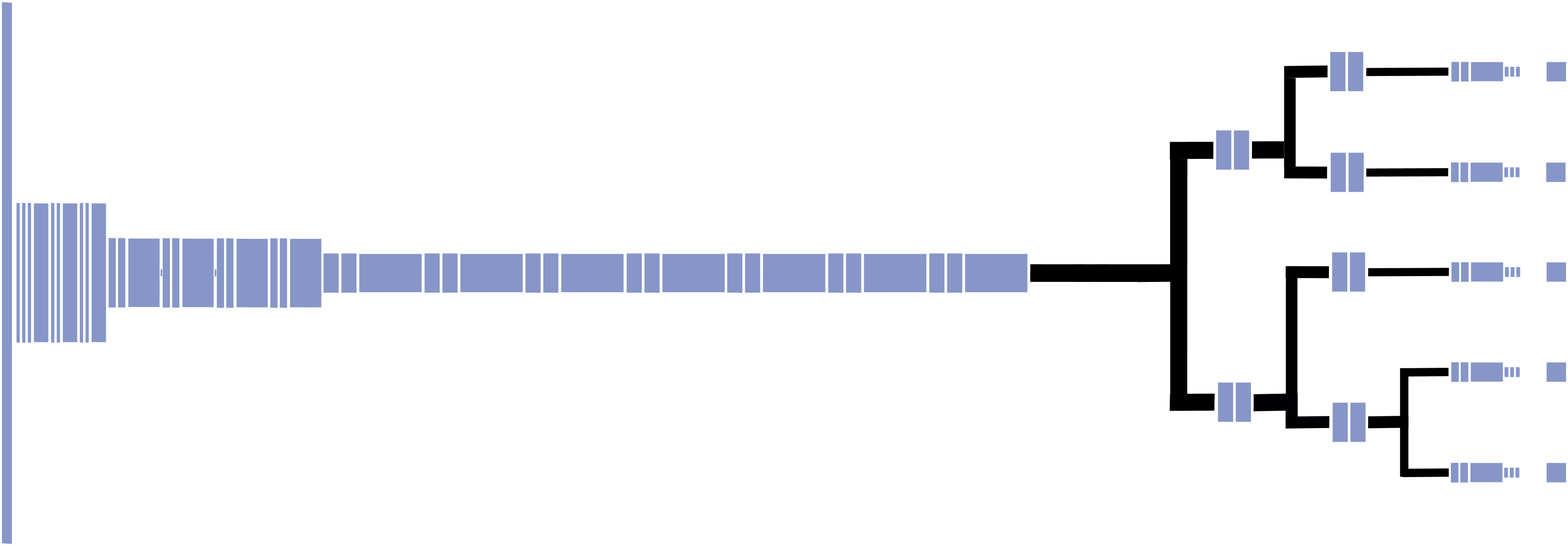}}}
    \caption{Hierarchical (NHL)}
    \label{fig:3a}
  \end{subfigure}
    \hfill
  \begin{subfigure}[b]{0.3\textwidth}
    \fbox{%
    \reflectbox{\includegraphics[clip, trim=19.7cm 19.7cm 19.7cm 19.7cm, width=\textwidth, angle=180]   {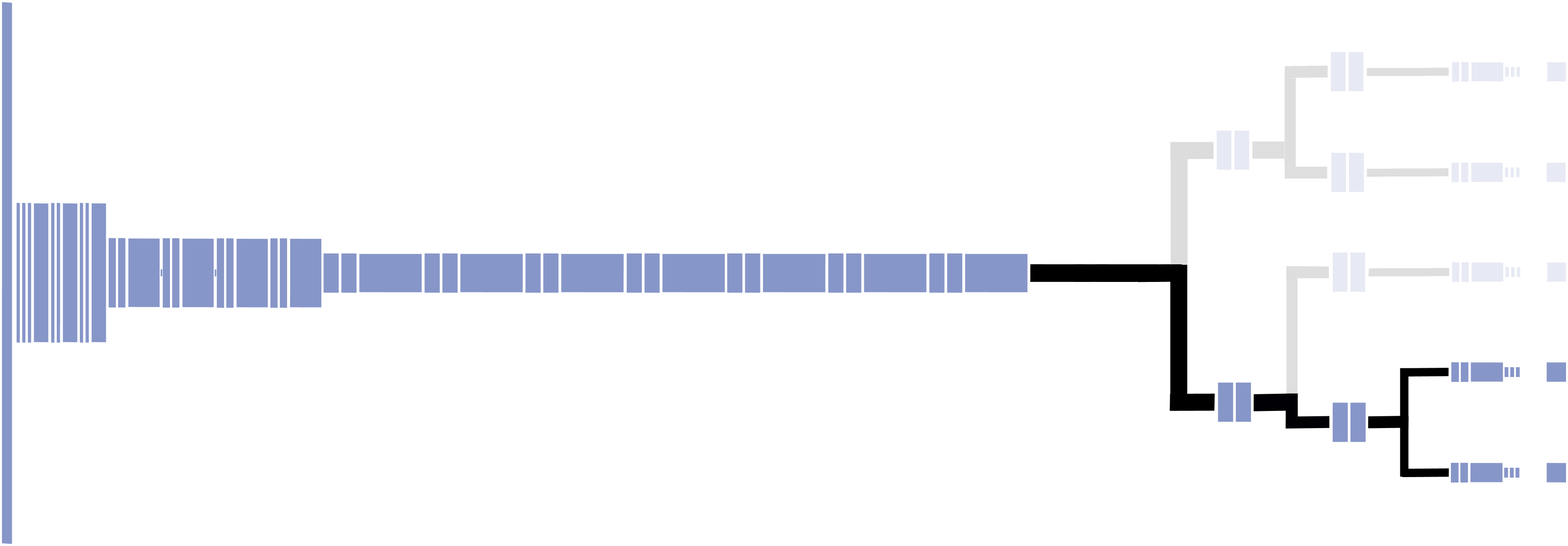}}}
    \caption{Efficient expert}
    \label{fig:4a}
  \end{subfigure} 
  \caption{Neural network topologies (blue: tensors, black: connection path). \subref{fig:1a} conventional deep network from left to right processing (the dots represent five categories); 
   \protect\subref{fig:3a} Our proposed nested structure; \protect\subref{fig:4a} Our innovative expert extraction technique.}
  \label{Main figure}
\end{figure}
Integration of the hierarchy in a classic CNN (c.f. \autoref{fig:1a}): The last conv block, the fifth in ResNet50 and ConvNeXtV2, is deleted and the FC is removed. 
The first branching point for the hierarchy is followed, the 'split point'. This is the first of three branching options.
We construct the \textbf{N}est features at a \textbf{H}igh-\textbf{L}evel (\textbf{NHL}) hierarchy (c.f. \ref{fig:3a}) by estimating the similarity between the categories using a quick semi-automatic approach with OpenAI’s ChatGPT4 \citep{chatgpt3} for all ImageNet's 1k categories; a single example:
1) marine animals, 2) sharks, 3) hammerhead shark.
Channels decrease with each new hierarchy level: 
1) 128, 2) 64, and 3) 32; while maintaining the bottleneck structure. 
Each branch end has its own FC, which contributes a value per category to softmax-log-loss, so training proceeds as for many other classification tasks.
Once trained our proposal can be utilized for any combination of the learned categories, here named as expert.
ImageNet100 provides $\bm{2^{100}-1}$ and ImageNet1k $\bm{2^{1000}-1}$ possible experts (c.f. \ref{fig:4a}), which do not require retraining.

\section{Experiments}
To show our method's flexibility, we used a ResNet50 \citep{ResNet}, trained from scratch, and a pre-trained ConvNeXtV2 \citep{Woo_2023_CVPR}.
The ResNet50$_\mathbf{NHL}$ expert for 20 categories shows in \autoref{tab:detailed_comparison_transposed} that despite similar GMACs and 41.3 \% fewer parameters, it has a 2.8\% top-1 acc. improvement over baselines. 
Although the NHL with 1000 categories has become a bit large (in parallel), 
its expert with 5 categories shows the same top-1 acc. as a deeper, pre-trained and fine-tuned ConvNeXtV2, despite 13.2\% fewer GMACs and 51.4\% fewer parameters. 
Such task-specific adjustments demonstrate the adaptability of our method over baselines that are not applicable (N/A) in this a way.
\begin{table}[htbp]
\centering
\caption{Comparison of ResNet50 and ConvNeXtV2 with our NHL and 2 exemplary experts.}
\label{tab:detailed_comparison_transposed}
{\small 
\begin{tabularx}{\textwidth}{>{\hsize=1.2\hsize}X | >{\centering\arraybackslash\hsize=0.9\hsize}X | >{\centering\arraybackslash\hsize=0.9\hsize}X | >{\centering\arraybackslash\hsize=0.9\hsize}X | >{\centering\arraybackslash\hsize=0.9\hsize}X}
\toprule
 & \multicolumn{2}{c|}{\textbf{ImageNet100}} & \multicolumn{2}{c}{\textbf{ImageNet1k}} \\
\midrule
\textbf{Metric} & \textbf{ResNet50} & \textbf{ResNet50$_\mathbf{NHL}$} & \textbf{ConvNeXtV2} &\textbf{ConvNeXtV2$_\mathbf{NHL}$}\\
\midrule
Top-1 acc. & 85.3\% & \textbf{85.7\%} & 80.1\% & \textbf{81.1\%} \\
GMACs & \textbf{4.13} & 5.86 & \textbf{4.46} & 14.55 \\
GMACs of expert & N/A & 4.22 & N/A  & \textbf{3.87} \\
GMACs reduction & 0\% & \textbf{-28\%} & 0\% & \textbf{-73.4\%} \\
Parameter & \textbf{23.5 / 23.7M} & 25M & \textbf{27.8 / 28.6M} & 119.9M \\
Parameter of expert & N/A  & \textbf{13.8M} & N/A  & \textbf{13.5M} \\
Parameter reduction & 0\% & \textbf{-44.5\%} & 0\% & \textbf{-88.7\%} \\
Train categories & 20 / 100 & 100 & 5 / 1000 & 1000 \\
Val categories & 20 & 20 & 5 & 5 \\
Top-1 acc. & 88.5 / 90.6\% & \textbf{93.4\%} & \textbf{80.0} / 79.6\% & \textbf{80.0\% }\\
\bottomrule
\end{tabularx}
}
\end{table}
\section{Conclusion}
We introduce a novel principle for arbitrarily matching computational complexity to task complexity without compromising predictive performance. This method is especially promising for mobile computing, industrial, drone, robotics, and edge device applications, 
where very specific tasks are defined such that the relevant high-level features can be dynamically selected as needed, utilizing our method.
Our experts have already achieved better top-1 accuracy with similar GMACs or the same top-1 accuracy with fewer GMACs, but we are confident that optimizing the base NHL model will unlock even more potential for low-resource computing.

\vspace{12pt}
\noindent
\emph{Acknowledgement:} Funded by the Deutsche Forschungsgemeinschaft (DFG, German Research Foundation) in the project Crossmodal Learning, TRR 169.

\bibliography{iclr2024_conference}
\bibliographystyle{iclr2024_conference}


\end{document}